\documentclass{article}

\usepackage{multirow}
\usepackage{arxiv}
\usepackage[utf8]{inputenc} 
\usepackage[T1]{fontenc}    
\usepackage{hyperref}       
\usepackage{url}            
\usepackage{booktabs}       
\usepackage{amsfonts}       
\usepackage{nicefrac}       
\usepackage{microtype}      
\usepackage{lipsum}
\usepackage{microtype}
\usepackage{amsmath}
\usepackage{booktabs}
\usepackage{graphicx}
\graphicspath{ {./images/} }

\title{Health Disparities through Generative AI Models: A Comparison Study Using 
 A Domain Specific large language model}

\author{
 Yohn Jairo Parra Bautista, Ph.D. \\
  Computing Information Science Department\\
  Florida A\&M University\\
  Tallahassee, FL 32317 \\
  \texttt{yohn.parrabautista@famu.edu} \\
   \And
 Vinicious Lima\\
  Computer Graphics Department\\
  Purdue University\\
  West Lafayette, IN 47906 \\
  \texttt{vlima@purdue.edu} \\
  \And
 Carlos Theran, PhD \\
  Computing Information Science Department\\
  Florida A\&M University\\
  Tallahassee, FL 32317 \\
  \texttt{carlos.theran@famu.edu} \\
  \And
 Richard Aló, PhD \\
  College of Science and Technology\\
  Florida A\&M University\\
  Tallahassee, FL 32317 \\
  \texttt{carlos.theran@famu.edu} \\
}

\begin{document}
\maketitle
\begin{abstract}
Health disparities are differences in health outcomes and access to healthcare between different groups, including racial and ethnic minorities, low-income people, and rural residents. An artificial intelligence (AI) program called large language models (LLMs) can understand and generate human language, improving health communication and reducing health disparities. There are many challenges in using LLMs in human-doctor interaction, including the need for diverse and representative data, privacy concerns, and collaboration between healthcare providers and technology experts. We introduce the comparative investigation of domain-specific large language models such as SciBERT with a multi-purpose LLMs BERT. We used cosine similarity to analyze text queries about health disparities in exam rooms when factors such as race are used alone. Using text queries, SciBERT fails  when it doesn't differentiate between queries text: "race" alone and "perpetuates health disparities." We believe clinicians can use generative AI to create a draft response when communicating asynchronously with patients. However, careful attention must be paid to ensure they are developed and implemented ethically and equitably. 
\end{abstract}


\section{Introduction}

\subsection{Addressing Health Disparities Through AI Models: An Important Issue}

Healthcare research and practice have long been marked by inequalities in health outcomes and difficulty accessing healthcare services. These discrepancies are called health disparities and bring great concern with using artificial intelligence (AI)\cite{gehlert2010importance}. Some efforts to reduce these disparities have become a vital public health priority, and healthcare professionals are increasingly exploring using artificial intelligence models as a potential solution to improve healthcare outcomes.

Healthcare professionals are using artificial intelligence models more frequently to combat various health disparities, such as those resulting from differences in race, ethnicity, socioeconomic status, and geographical location \cite{zou2021ensuring}. These models can potentially analyze vast amounts of data, recognize patterns and trends, and offer individualized and specific interventions to improve health outcomes for all groups \cite{han2020bridging}.

Health disparities, which refer to differences in health outcomes among different population groups, remain a significant challenge in healthcare. Generative artificial intelligence (AI) models are emerging as promising approaches to address these disparities \cite{harrer2023attention}. Generative AI models can analyze vast amounts of data, identify patterns and trends, and generate new data, such as text, images, and audio. The potential of these models to provide personalized and targeted interventions to improve health outcomes for all populations has captured the interest of healthcare researchers and practitioners \cite{arora2023promise}. 

A generative adversarial network (GAN) is an AI model that generates synthetic medical images to tackle racial disparities in dermatology. A particular study revealed that GAN effectively produced synthetic images that were clinically relevant and accurately represented the underlying pathology  \cite{saaran2021literature}. This approach can potentially improve access to dermatological care for underserved populations.

Another example is a deep learning model that identifies and mitigates disparities in cervical cancer screening rates \cite{tan2021automatic}. The study demonstrated that the model could predict which patients were at increased risk of missing screening appointments and provided targeted interventions to increase screening rates.

In terms of the potential of natural language processing (NLP) models in enhancing health equity through the analysis of electronic health records (EHRs) \cite{dreisbach2019systematic}, research has shown promising results. By analyzing EHRs, NLP models can identify disparities in health outcomes and assist in developing targeted interventions for underserved populations. However, it is crucial to consider social determinants of health, such as race, ethnicity, and socioeconomic status, when designing and implementing NLP models to address health disparities\cite{koleck2019natural}. Using domain-specific language models tailored to specific healthcare domains can also improve the accuracy and effectiveness of NLP models in addressing health disparities. In general, NLP models have the potential to have a significant impact in promoting health equity and reducing health disparities in healthcare. 

The primary objective of this comparative study is to investigate the application of domain-specific large language models such as SciBERT in healthcare research to address health disparities through generative AI models \cite{ghassemi2019practical}. The study aims to offer an assessment of two models to provide valuable insights into the potential of AI models in resolving health disparities. SciBERT has about 3x as many tokens with digits. SciBERT's number of tokens is much more diverse. They are often subwords, and many include decimal places or other symbols like "\%" or "(.".

In this study, we analyze text queries that involve health disparities. We use cosine similarity to test two LLMs: SciBERT and BERT, on those queries. We further test the cosine similarity in three scenarios where domain-specific LLMs like SciBERT do not make a distinction on text query that involved health disparities.

Cosine similarity is a widely used metric to measure the similarity between two vectors in a high-dimensional space\cite{jairo2019comparison,luo2018cosine}. In language models, we often represent words or documents as vectors in a vector space, where each dimension corresponds to a feature or a characteristic of the text.

The formula for cosine similarity:

\[
\text{similarity}(u, v) = \frac{u \cdot v}{\lVert u \rVert \lVert v \rVert}
\]

- $u \cdot v$ is the dot product between the two vectors, which is the sum of the element-wise products of the vectors.
- $\lVert u \rVert$ and $\lVert v \rVert$ are the Euclidean norms of the vectors, which are the square roots of the sum of the squares of the vector elements.

Cosine similarity is commonly employed in language models to assess the similarity between two documents, frequently represented as vectors of word frequencies or embeddings. For example, suppose we have two documents $d_1$ and $d_2$ represented as vectors of word frequencies or embeddings. We can calculate their cosine similarity as:

$$\text{similarity}(d_1, d_2) = \frac{\sum_{i=1}^{n} w_{1,i} \times w_{2,i}}{\sqrt{\sum_{i=1}^{n} w_{1,i}^2} \times \sqrt{\sum_{i=1}^{n} w_{2,i}^2}}$$

where $n$ is the number of words in the vocabulary, $w_{1,i}$ and $w_{2,i}$ are the word frequencies or embeddings of the $i$-th word in the two documents $d_1$ and $d_2$, respectively.

This formula measures the cosine of the angle between the two vectors, ranging from -1 (opposite directions) to 1 (the same direction). A value of 0 indicates that the two vectors are orthogonal (perpendicular) to each other, meaning they have no similarity.

\section{Literature Review}
\subsection{Understanding Health Disparities through AI: Previous studies}

Over the past few decades, machine learning has emerged as a game-changing technology with the potential to revolutionize various domains, including healthcare. In the medical field, artificial intelligence/machine learning (AI/ML) has made significant advances in the diagnosis and prognosis of diseases, allowing researchers to understand the behavior of the human body better and classify medical images or videos \cite{bohr2020rise}. One of the most recent and notable applications of machine learning in healthcare is its ability to accurately detect COVID-19 by analyzing chest radiographs, reducing the risk of human transmission and infection. With the development of generative adversarial networks (GANs) and other machine learning algorithms, the potential for advancements in the healthcare sector is vast. The possibilities will continue to grow as technology advances\cite{jamshidi2020artificial}.

Artificial intelligence (AI) in healthcare care is a promising field that could revolutionize medicine by enabling more accurate decisions and personalized treatment. However, progress in this area is limited by legal and ethical issues, such as patient data privacy and concerns about non-consensual data sharing \cite{han2020bridging}. Researchers have proposed a novel framework for generating synthetic data that closely approximates variables' joint distribution in an original electronic health record (EHR) dataset to address this. Using a conditional generative adversarial network (GAN) framework, the proposed model, GAN, demonstrated reliability in joint distributions and consistently outperformed state-of-the-art methods in evaluating similarity in model performance across four independent datasets \cite{kather2022medical}. This method could provide a solution to more open data sharing and enable the development of AI solutions while lowering the risk of breaching patient confidentiality.

A potential gap in the GAN model is the lack of discussion on the effectiveness of the proposed GAN method compared to accurate patient data. While the model was evaluated for similarity in model performance across four independent datasets, it is unclear how well the generated synthetic data corresponds to the actual patient data. Further research could address this gap and assess the practical applicability of the GAN framework in real-world healthcare settings.

AI in healthcare and medicine has become increasingly popular. Still, the quality and diversity of training data are critical in ensuring that the algorithms function effectively and are free of biases. Biases can arise from sample selection biases and class imbalances, which can degrade the performance of trained AI models \cite{chen2021synthetic}. Synthetic data, which can be created using accurate forward models, physical simulations, or AI-driven generative models, have been proposed as a potential solution to overcome the lack of annotated medical data in real-world settings. However, using synthetic data raises several issues, such as their possible misuse for malicious purposes and the need for regulatory frameworks to govern their usage in modifying AI algorithms in healthcare \cite{torfi2020corgan}. Despite these challenges, synthetic data can increase the diversity and representativeness of datasets and improve the robustness and adaptability of AI models, leading to better medical decisions in a broader range of real-world environments. Further research is needed to fully understand the benefits and limitations of synthetic data in healthcare and to develop appropriate policies to govern their use.

Studies have shown that healthcare providers' implicit biases can influence their clinical decision-making, communication, and treatment recommendations, leading to differential treatment and outcomes for patients of different races, ethnicities, sex, and socioeconomic backgrounds. Implicit bias, which refers to unconscious attitudes or stereotypes that can affect an individual's behavior, has been recognized as a significant factor contributing to health disparities in medical care \cite{hoffman2016racial}. For example, a provider may use words such as "non-compliant" or "difficult" to describe a patient who misses appointments or does not follow their recommended treatment plan, which can reflect negative stereotypes about specific populations. Additionally, healthcare providers may use language that assumes a patient's cultural background or beliefs, such as assuming that a Hispanic patient is a migrant worker or that an Asian patient practices traditional medicine.

\subsubsection{Domain-Specific Large Language Models for Healthcare}

Recent advances in pre-trained contextualized language models (PLMs) have developed several domain-specific PLMs, enabling various downstream applications \cite{naseem2022benchmarking}. In mental healthcare, the need for existing PLMs has hindered the potential benefits of such technology. However, with the release of MentalBERT and MentalRoBERTa, two pre-trained masked language models for mental healthcare, the research community now has access to tools that can benefit the early detection of mental disorders and suicidal ideation \cite{ji2021mentalbert}. Similarly, social media has been recognized as a potential source of information for public health surveillance (PHS). Despite the potential, the technology is still in its early stages, and currently, there are no PLMs for PHS-related social media tasks. The introduction of PHS-BERT, a transformer-based PLM for identifying PHS tasks on social media, opens up new possibilities for the community to monitor disease trends and emergency cases, track disease awareness and response, and identify possible outbreaks\cite{naseem2022benchmarking}. By comparing and benchmarking the performance of PHS-BERT on 25 datasets from different social media platforms related to seven other PHS tasks, the research community can reduce computational costs and introduce new baselines for future work across various PHS-related tasks. Overall, the development and evaluation of domain-specific PLMs such as MentalBERT, MentalRoBERTa, and PHS-BERT demonstrate the potential of PLMs to benefit various downstream applications and encourage further research in developing domain-specific PLMs.

Using general-purpose language models in natural language processing for downstream modeling has become increasingly popular, but the text embeddings they generate can pose privacy risks\cite{zheng2022pretrained}. For instance, the privacy risks of eight state-of-the-art language models, including Google's Bert and OpenAI's GPT-2, demonstrate how sensitive information can be reverse-engineered from unprotected embeddings. A study proposes four defenses to obfuscate these embeddings and mitigate privacy risks. The study's findings have significant implications for the future use of general-purpose language models in sensitive data applications and highlight the need for further research into mitigation techniques that balance privacy and utility \cite{brinkmann2021improving}.

BERT (Bidirectional Encoder Representations from Transformers) is a powerful and popular pre-trained language model developed by Google. BERT has significantly advanced the natural language processing (NLP) field by achieving state-of-the-art results in various downstream NLP tasks, such as question answering, sentiment analysis, and text classification \cite{devlin2016bert}. BERT employs a transformer-based architecture, which allows it to capture complex relationships between words and produce context-aware embeddings that represent the underlying meaning of a sentence or document. Additionally, BERT can be fine-tuned on specific downstream tasks by adding task-specific layers on top of the pre-trained model, improving its performance on those tasks \cite{ettinger2020bert}. BERT's success has led to widespread adoption in industry and academia and has opened up new possibilities for developing advanced NLP applications. However, the large size of BERT and its resource-intensive nature pose challenges for practical applications on resource-constrained devices. Therefore, researchers actively explore ways to compress and optimize BERT models for more efficient deployment in real-world applications.

SciBERT is a pre-trained language model developed by the Allen Institute for Artificial Intelligence (AI2) in collaboration with the University of Washington and the Allen Institute for Cell Science. Unlike BERT, which was trained on a general corpus of text, SciBERT is specifically designed to improve the performance of natural language processing (NLP) models on scientific and biomedical texts\cite{beltagy2019scibert}. SciBERT is trained in a large corpus of scientific articles, including papers from biology, chemistry, and computer science. The model also includes specialized vocabularies and embeddings to represent better the complex language and terminology used in scientific publications \cite{gu2021domain}. These features have been shown to significantly improve the performance of SciBERT on a range of scientific NLP tasks, such as named entity recognition, relation extraction, and sentence classification. In addition, SciBERT's pre-training can be fine-tuned on specific downstream tasks, allowing it to achieve state-of-the-art performance on a range of biomedical and scientific tasks\cite{usha2022named}. The success of SciBERT has led to increased interest in developing domain-specific language models that can better capture the nuances of specialized domains, opening up new possibilities for developing advanced NLP applications in various scientific and technical fields.

In a recent research article, ChatGPT was evaluated in the context of improving knowledge about healthcare topics, such as vaccinations and lifestyle choices. The article highlighted that although ChatGPT has the potential to be a valuable tool in this domain, concerns remain regarding the reliability of the information it provides. Specifically, it is unclear whether the source of information used by ChatGPT can be considered trustworthy and accurate. This raises important questions regarding the use of ChatGPT in healthcare contexts and the need for further research to ensure that the information it provides is reliable and of high quality.\cite{biswas2023role}. ChatGPT is a language model based on GPT3, so it can potentially perpetuate bias based on its training data \cite{borji2023categorical}. Indeed, developers of ChatGPT modify or filter the training data to reduce bias in its outputs. However, the specific details of this process still need to be fully disclosed by OpenAI. It is important to note that, while debiasing is crucial, it can also mask reality, and the precision of using such a model to reveal bias issues may be questioned \cite{ferrara2023should}.

\section{Methodology}

Health disparities are a critical issue affecting the well-being of various groups, including low-income populations, rural communities, and racial and ethnic minorities \cite{bailey2017structural}. Addressing these disparities is crucial to ensuring equitable access to healthcare and reducing health inequalities. Large language models (LLMs) have emerged as a promising technology to aid this effort. By enabling natural language processing and generation, LLMs have the potential to improve health communication and increase productivity in the exam room. However, their use in healthcare also presents challenges, such as ensuring data diversity and addressing privacy and bias concerns\cite{simpao2014review}. This study highlights the potential of LLMs, specifically, BERT and SciBERT, to improve physician productivity in the exam room. However, it also demonstrates the need for careful attention to the ethical and equitable development and implementation of these technologies. As we continue to explore the use of LLMs in healthcare care, it is critical to prioritize the needs of underserved populations and ensure that these technologies do not perpetuate health disparities.

\begin{figure}[ht!]
\centering
\includegraphics[width=1.0\textwidth,height=0.60\textheight]{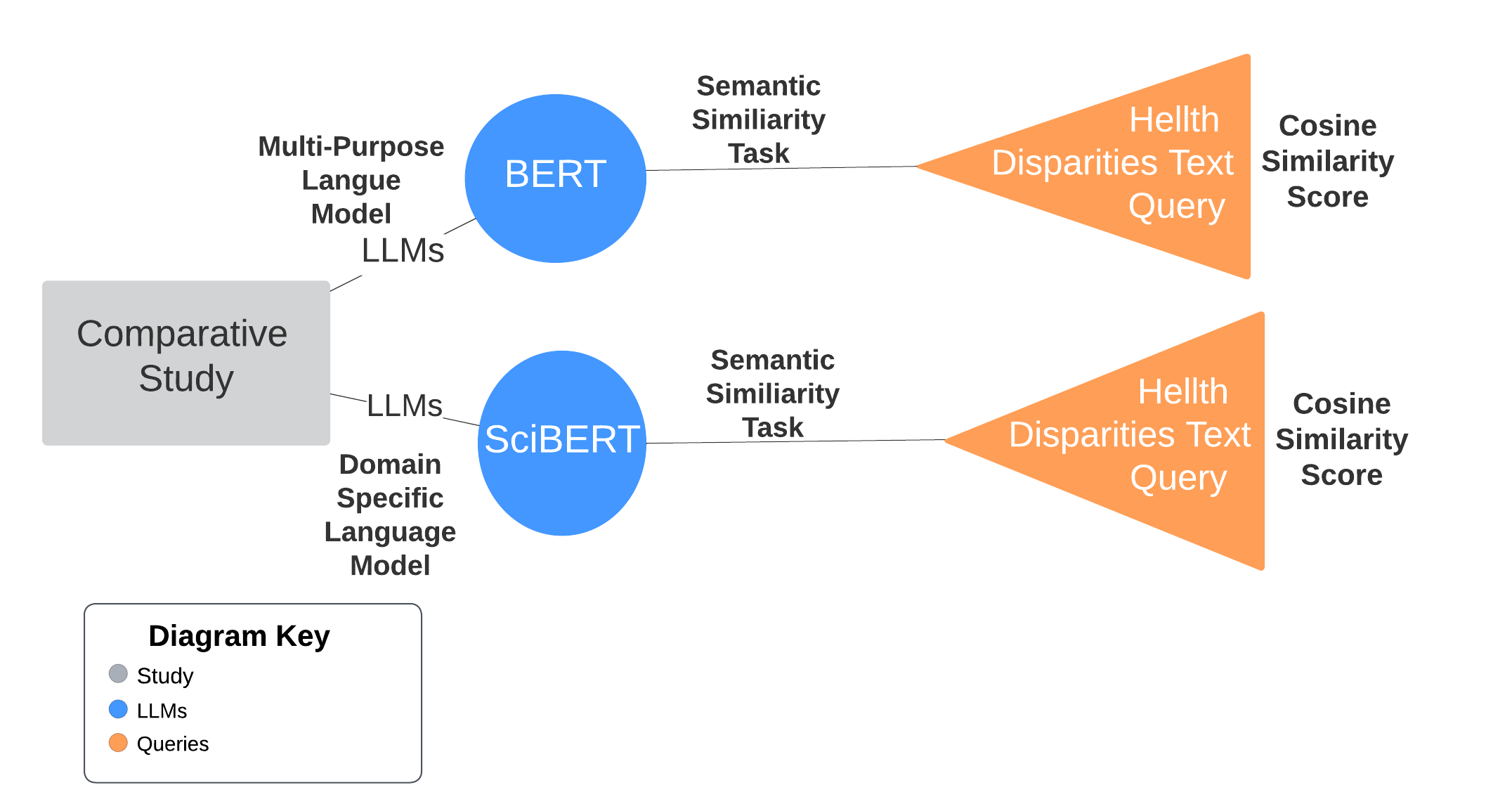}
\caption{Flowchart Comparative Study with Health Disparities prompts}
\label{fig:flowchart-framework} 
\end{figure}

Researchers have developed domain-specific language models by training the BERT architecture from scratch on a corpus specific to a particular domain instead of using the general-purpose English text corpus utilized to train the original BERT model. This methodology results in a language model with vocabulary and word embeddings more tailored to the needs of domain-specific NLP tasks than the original BERT model. A few instances of such models include:

SciBERT (biomedical and computer science literature corpus)

FinBERT (financial services corpus)

BioBERT (biomedical literature corpus)

ClinicalBERT (clinical notes corpus)

mBERT (corpora from multiple languages)

patentBERT (patent corpus)

We used BERT and SciBERT to test cosine similarity in different contexts of health disparities of text queries. Research on potential biases in AI toward healthcare decisions is currently insufficient. Figure 1 shows the method used to text similarities using BERT and SciBERT. Our approach uses a semantic similarity task; we have three pieces of text, calling them "query", "A", and "B", which are all on health disparities topics. We did this structure so that the query text is always more similar to A than B.

In NLP, words are often represented as numerical vectors in a high-dimensional space called an embedding space, also known as word embeddings. Word embeddings capture the meaning of words in the context of the language they are used in and enable computational models to reason about the meaning of text. One common way to measure the similarity between two word embeddings is to use cosine similarity. Cosine similarity calculates the cosine of the angle between two vectors in the embedding space and ranges from -1 (opposite directions) to 1 (same direction). To calculate cosine similarity, we take the dot product of the two vectors and divide it by the product of their lengths. 

\section{Results}

\begin{table}[ht!]
\centering
\renewcommand{\arraystretch}{1.5} 
\begin{tabular}{@{}p{3.5cm}p{9.5cm}cc@{}} 
\toprule
\textbf{Labels} & \multicolumn{1}{c}{\textbf{Target Text}} & \textbf{Models} & \textbf{CS} \\
\midrule
\textbf{Text Query 1} & During a routine check-up, a patient from a historically marginalized community discussed their struggles with accessing affordable healthcare, highlighting health disparities in the doctor's room. & & \\
\midrule
\multirow{2}{3.5cm}{Query A} & \multirow{2}{9.5cm}{The doctor expressed concern over the significant health disparities low-income communities face.} & SciBERT & \textbf{0.91} \\
& & BERT & 0.86 \\
\midrule
\multirow{2}{3.5cm}{Query B} & \multirow{2}{9.5cm}{The observed cosmic shear auto-power spectrum receives an additional contribution due to shape noise from intrinsic galaxies.} & SciBERT & \textbf{0.83} \\
& & BERT & 0.60 \\
\bottomrule
\end{tabular}
\caption{Cosine similarity score with health disparities against astrophysics topic.}
\label{tab:table1}
\end{table}

\begin{table}[ht!]
\centering
\renewcommand{\arraystretch}{1.5} 
\begin{tabular}{@{}p{3.5cm}p{9.5cm}cc@{}} 
\toprule
\textbf{Labels} & \multicolumn{1}{c}{\textbf{Target Text}} & \textbf{Models} & \textbf{CS} \\
\midrule
\textbf{Text Query 2} & Food desert is an area with little to no access to affordable and nutritious food, which can lead to poor health outcomes for residents. For instance, in the rural community of Meadowville, there is no grocery store within a 50-mile radius. & & \\
\midrule
\multirow{2}{3.5cm}{Query A} & \multirow{2}{9.5cm}{Food deserts have been linked to health disparities, such as higher rates of obesity, diabetes, and cardiovascular disease.} & SciBERT & \textbf{0.92} \\
& & BERT & 0.81 \\
\midrule
\multirow{2}{3.5cm}{Query B} & \multirow{2}{9.5cm}{Life expectancy refers to the average years a person is expected to live based on their birth year, gender, and other characteristics.} & SciBERT & \textbf{0.93} \\
& & BERT & 0.76 \\
\bottomrule
\end{tabular}
\caption{Cosine similarity score with food desert against life expectancy topic.}
\label{tab:table2}
\end{table}

\begin{table}[ht!]
\centering
\renewcommand{\arraystretch}{1.5} 
\begin{tabular}{@{}p{3.5cm}p{9.5cm}cc@{}} 
\toprule
\textbf{Labels} & \multicolumn{1}{c}{\textbf{Target Text}} & \textbf{Models} & \textbf{CS} \\
\midrule
\textbf{Text Query 3} & Race can indicate certain health risks, such as sickle cell anemia in African descent or Tay-Sachs disease in individuals of Ashkenazi Jewish descent. & & \\
\midrule
\multirow{2}{3.5cm}{Query A} & \multirow{2}{9.5cm}{AI has the potential to perpetuate health disparities if not developed and used responsibly.} & SciBERT & \textbf{0.96} \\
& & BERT & 0.91 \\
\midrule
\multirow{2}{3.5cm}{Query B} & \multirow{2}{9.5cm}{AI has the potential to help reduce health disparities by improving access to quality healthcare.} & SciBERT & \textbf{0.96} \\
& & BERT & 0.90 \\
\bottomrule
\end{tabular}
\caption{Cosine similarity score: AI improving health disparities using race alone vs. AI perpetuating health disparities topic}
\label{tab:table3}
\end{table}

\begin{table}[ht!]
\centering
\renewcommand{\arraystretch}{1.5} 
\begin{tabular}{@{}p{3.5cm}p{9.5cm}cc@{}} 
\toprule
\textbf{Labels} & \multicolumn{1}{c}{\textbf{Target Text}} & \textbf{Models} & \textbf{CS} \\
\midrule
\textbf{Text Query 4} & Race, external factors, diet, and age can indicate certain health risks, such as sickle cell anemia in African descent or Tay-Sachs disease in individuals of Ashkenazi Jewish descent. & & \\
\midrule
\multirow{2}{3.5cm}{Query A} & \multirow{2}{9.5cm}{AI has the potential to perpetuate health disparities if not developed and used responsibly.} & SciBERT & \textbf{0.90} \\
& & BERT & 0.74 \\
\midrule
\multirow{2}{3.5cm}{Query B} & \multirow{2}{9.5cm}{AI has the potential to help reduce health disparities by improving access to quality healthcare.} & SciBERT & \textbf{0.88} \\
& & BERT & 0.71 \\
\bottomrule
\end{tabular}
\caption{Cosine similarity score: AI improving health disparities using race plus other factors vs. AI perpetuating health disparities topic}
\label{tab:table3}
\end{table}

SciBERT is a variant of BERT specifically trained on scientific papers, making it a useful tool for natural language processing (NLP) tasks in the scientific domain \cite{arora2021innovators}. Several studies have demonstrated the effectiveness of SciBERT in various NLP tasks, such as extracting scientific information from images and classifying drugs based on free data \cite{patricoski2022evaluation}. Additionally, SciBERT has been shown to outperform the original BERT model in assigning higher similarity scores to matched clinical trial texts.

Cosine similarity has been used in various tasks, such as text classification and question answering \cite{sidorov2014soft}. For example, in a study on the scientific literature on COVID-19, SciBERT was used as a pointer network to select an answer start and end index given a question and a paragraph \cite{otegi2020automatic}. In another study, researchers proposed a method for determining the similarity of text documents for the Kazakh language, taking into account synonyms, and evaluated the method's effectiveness using cosine, Dice, and Jaccard similarity measures \cite{bakiyev2022method}.

We used the first two queries in Query 1 and 2 as a baseline, shown in Tables 1 and 2. We found that SciBERT correctly distinguishes for query B in both cases. However, in Query 3, Table 3, SciBERT does not distinguish nither to query A or B. Instead, BERT was slightly biased toward query A. When we add other factors besides race in Query 4, Table 4, which improves the similarity towards Query A.

We compare two models for semantic similarity and how well they rank the similarities. We did not compare the specific cosine similarity values across the two models. We have structured our example as "is query more similar to A or to B?"

Query three and A discuss race as an indicator in a certain disease, such as sickle cell anemia, and how the race factor can perpetuate health disparities. In contrast, B discusses the positive side of AI in health disparities. However, to recognize the similarity between queries three and B, LLMs must know that race alone is not responsible for anemia in African descent.

The data analysis was performed using Python software (Google Colab), and the code used for the analysis is available in the following GitHub repository: 

\url{https://github.com/Yohnjparra/CosineSimilarity-HealthDisparities/blob/main/Generative_AI_Health_Disparities.ipynb}

We believe that SciBERT, trained in biomedical text, would better distinguish similarities in a healthcare disparities context than BERT. However, BERT, a multipurpose language with a lower similarity score, makes a slight distinction better. 

Cosine similarity is a widely used measure for comparing the similarity between two vectors in a vector space model (VSM). However, some studies have pointed out that cosine similarity can be biased by features of higher values and may not consider how many features two vectors share. To address these issues, researchers have proposed several modifications to cosine similarity, such as the soft cosine measure and the distance-weighted cosine similarity measure.

In our study, we examine the performance of BERT and SciBERT in recognizing the role of race as a factor in certain diseases. We found that in some cases, the models were biased toward race as a unique factor, which can perpetuate health disparities. Specifically, we identified a few examples where "race" as a unique factor is already biased in the trained corpus data for BERT and SciBERT. This highlights the importance of carefully examining the training data and ensuring they are representative and unbiased. Additionally, it underscores the need for ongoing efforts to improve the fairness and accuracy of NLP models, particularly in the context of health disparities. Our findings suggest that further research is needed to develop more effective methods for addressing bias in NLP models and promoting health equity.

\section{Discussion}

Health disparities are a critical issue that affects the well-being of various social groups, including low-income populations, rural communities, and racial and ethnic minorities. Addressing these disparities is crucial to ensuring equitable access to healthcare and reducing health inequalities. Large language models (LLMs) have emerged as promising technology that can potentially aid this effort. By enabling natural language processing and generation, LLMs have the potential to improve health communication and increase productivity in the exam room. However, its use in healthcare also presents challenges, such as ensuring data diversity and addressing privacy and bias concerns. This study highlights the potential of LLMs, specifically, BERT and SciBERT, to improve physician productivity in the exam room. However, it also demonstrates the need for careful attention to the ethical and equitable development and implementation of these technologies. As we continue to explore the use of LLMs in healthcare care, it is critical to prioritize the needs of underserved populations and ensure that these technologies do not perpetuate health disparities.

Integrating Large Language Models (LLMs) into healthcare workflows can significantly increase the productivity of physicians and other healthcare workers, from back-office staff to schedulers. Using generative AI to create draft responses when communicating asynchronously with patients can save time and improve the quality of care. However, it is essential to ensure that the LLMs are appropriately trained on healthcare terminology and do not perpetuate healthcare disparities. There is a risk of increasing healthcare inequalities when LLMs do not accurately identify similarities between sentences or words in healthcare terminology. It is crucial to recognize that LLMs are not a panacea and that their integration must be approached with caution and thoughtful consideration of ethical and equitable implications. With proper implementation, LLMs can be valuable in reducing healthcare disparities and improving access to care for underserved populations.

Our results demonstrate that BERT outperformed SciBERT when comparing text queries on health disparities. This result is somewhat surprising given that SciBERT was specifically designed as a domain-specific model trained in biomedical corpora. This suggests that disparities may not have been adequately represented in the training data for SciBERT, whereas BERT has a broader background training across diverse corpora. One can also suggest that data used as the basis for SciBERT give low importance on ethical issues compared to the core scientific information.

\section{Conclusions}

In conclusion, health disparities are a pressing issue that affects various social groups and must be addressed to ensure equitable access to health care and reduce health inequalities. Large language models (LLMs) hold significant potential for improving healthcare productivity and communication, but their implementation must be approached with care. The study highlights the potential of LLMs, such as BERT and SciBERT, to increase physician productivity in the exam room. However, it also emphasizes addressing ethical and equitable development and implementation, including data diversity, privacy, and bias concerns. To ensure that LLMs do not perpetuate health disparities, it is critical to prioritize the needs of underserved populations. Although LLMs are not a panacea, they can be valuable in reducing healthcare disparities and improving access to care with proper implementation.

Our goal is to demonstrate how some pre-trained language models may not effectively account for health disparity contexts. It is important to consider this topic, as language models are expanding in the health field, and ethical issues may arise if they are not properly applied. Health disparities are a critical concern, and it is crucial to ensure that language models are sensitive to these disparities to avoid exacerbating existing biases. By highlighting the limitations of some pre-trained language models in this regard, we hope to draw attention to the need for more inclusive and equitable approaches to natural language processing in healthcare.

\begin{center}
  ACKNOWLEDGMENT  
\end{center}

The authors acknowledge NIH BioMed Grant Number 150108136 under Florida A\&M University and CI-New: Cognitive Hardware and Software Ecosystem Community Infrastructure for allowing us to run our application in their infrastructure (Nautilus).

\bibliographystyle{unsrt}  
\bibliography{references}  






\end{document}